\documentclass{article}
\title{Named entity recognition using GPT for identifying comparable companies}
\author{Eurico Covas  \\ e-mail: \href{mailto:eurico.covas@mail.com}{eurico.covas@mail.com}\\
	Collaborator, Instituto de Astrof\'isica e Ci\^encias do Espa\c{c}o (IA), Portugal  \\
	}

\date{\today}

\usepackage{graphicx}
\usepackage[a4paper,margin=2.5cm,foot=2cm]{geometry}
\usepackage{booktabs,siunitx}
\usepackage[round]{natbib}   
\usepackage{courier} 
\usepackage{siunitx} 
\usepackage{hyperref}

\newenvironment{ttquotesmall}
 {\quote\ttfamily\raggedright\small}
 {\endquote}
 
\providecommand{\keywords}[1]
{
  \small	
  \textbf{\textit{Keywords ---}} #1
}
\begin{document}

\maketitle

\begin{abstract}
For both public and private firms, comparable companies' analysis is widely used as a method for company valuation. In particular, the method is of great value for valuation of private equity companies. The several approaches to the comparable companies' method usually rely on a qualitative approach to identifying similar peer companies, which tend to use established industry classification schemes and/or analyst intuition and knowledge. However, more quantitative methods have started being used in the literature and in the private equity industry, in particular, machine learning clustering, and natural language processing (NLP). For NLP methods, the process consists of extracting product entities from e.g., the company's website or company descriptions from some financial database system and then to perform similarity analysis. Here, using companies' descriptions/summaries from publicly available companies' Wikipedia websites, we show that using large language models (LLMs), such as GPT from OpenAI, has a much higher precision and success rate than using the standard named entity recognition (NER) methods which use manual annotation. We demonstrate quantitatively a higher precision rate, and show that, qualitatively, it can be used to create appropriate comparable companies peer groups which could then be used for equity valuation.
\end{abstract}


\section{Introduction}\label{introduction}

Company valuation is the process of attributing a value to a public or private company\footnote{Public companies' stocks are listed on an exchange, allowing for more transparency and scrutiny, while private companies' stocks are private held and a common realizable price for the shares is not easily or readily obtainable.}  in a certain currency and valuation date. For public companies, one could use e.g., the exchange share price multiplied by the total number of shares as a valuation\footnote{Given that share prices fluctuate short term in a stochastic fashion, one may want to take some averaging across time, as it is standard in the financial industry \citep[see e.g.,][]{huddart2009volume}.}, but often there is a speculative influence on the share price, making the stock price, at short time scales, more of a stochastic rather than a deterministic variable \citep[see e.g.,][]{black1986noise, fama1995random}. For private companies, one does not even have a publicly available share price, so several methods have been proposed for valuation \citep[see e.g.,][and references therein]{damodaran2012investment}. Nonetheless, all these quantitative methods can be used for any firm, regardless of being public or private. Three of the most   used methods \citep[see e.g., Table 3 in][for a survey of most commonly used methods]{https://doi.org/10.1111/1468-036X.00130} are respectively: the comparable companies or financial multiples / relative valuation method; the comparable transactions method; and the discounted cash-flow (DCF) method. These methods are used to value a company, rather than its stock, as these are mostly used for long term investment, or mergers and acquisitions (M\&A), rather than short-term trading or arbitrage. 

The first method, comparable companies, or relative valuation \citep[see e.g.,][]{liu2002equity}, is based on the ratios of one or some of the financial variables of the selected company to value against the ones from a peer group of public companies. These ratios are therefore dimensionless. As we can obtain the share value of the public companies from the stock market value, we have, therefore, the enterprise value of those companies, and through some kind of averaging (including possibly some statistical removal of outliers), we can obtain a peer or similar companies group ratio. Once the peer group ratios are obtained, one can use the financial statements of the selected company to do an inversion and obtain an enterprise value estimate for the company. The second method, comparable transactions \citep[see e.g.,][and references therein]{berkman2000accuracy}, analyses similar companies that have transacted recently, either an acquisition, or a merger (M\&A), or an initial public offer (IPO), or a seasoned equity offerings (SEO) or similar. Then, the same approach as the relative valuation is used, using available ratios based on the valuation related to the transaction. The third method, the discounted cash-flow method or DCF for short \citep[see e.g.,][and references therein]{inselbag1997two}, uses the future projected  cash-flows of the company, discounted to today, i.e., multiplied by a factor or factors to consider the future value of money. 
There are many other methods and quite a lot of variations on the above ones, but these are, according to some surveys \citep{liu2020information}, some of the most used approaches. Most of the equity valuation research (both public and private) is, however, done quite manually, whereby the equity researcher or financial analyst uses some data source of financial data (e.g., Bloomberg, Factset, Reuters -  now called Refinitiv, S\&P, UK Companies House, U.S. Securities and Exchange Commission) and a spreadsheet model\footnote{Sometimes the spreadsheet, usually in Excel, are augmented with some Visual Basic -- VBA code.} to value the selected company. This can lead to errors, biases, lack of traceability and reproducibility, etc. This is quite costly, and recently, there has been some advances in the  automation of this process, done by several fintech software houses, e.g., Valutico or Equity-X, in order to solve the problem of consistency and process industrialization.



For the first two valuation methods described above, the comparable companies and comparable transactions approaches, it is essential to define what we mean by similar or peer group companies. 
There are many other reasons to want to be able to identify comparable companies, e.g., for M\&A \citep{eaton2022peer}; for the owners to establishing the competitive landscape; for economic research into business networks; among others.
There are also many ways to identify the company's peer group, some approaches being qualitative, some being quantitative.
A particularly used method is to choose companies within the same sector by using a standard industry classification such as the
Standard Industrial Classification (SIC) system, the North American Industry Classification System (NAICS) and the
Global Industry Classifications Standard (GICS) system \citep{phillips2016industry}.
However as far back as \citet{king1966market} it has been noted that the industry grouping only contribute a small percentage of stock returns variance. Other studies \citep[see e.g.,][]{bhojraj2003s, hoberg2016text} have tried to show that most of these industry classifications (with maybe the exception of the GICS system), do not explain most of the cross-sectional share movement within a industry or sub-industry code.
  \citet{de2015analysts} found that research analysts or equity valuation experts tend to choose peer companies based on the size, asset turnover, the industry classification, and trading volume, among other less relevant variables. 
 Some quantitative approaches for peer selection, based on ``big data'' have also been proposed \citep{lee2015search,lee2016search}, whereby companies' peer groups are build based on top co-searches by financial analysts, using databases such as USA Electronic Data Gathering, Analysis and Retrieval system (EDGAR), part of the Security Exchange Commission (SEC), used by, e.g., sell-side stock analysts. The results seem to show that the so-called ``wisdom of the crowd'' approaches for peer construction can perform well. The results by \citet{kaustia2021common} seem to support this, that co-searches by analysts help to define more homogeneous peer groups than industry classifications. Some \citep{sprenger2011tweets} even used natural language processing (NLP) to extract data from social media (Twitter) to define a company network, i.e., peer groups that more readily explain stock movement than industry classification groups.
 \citet{knudsen2017stick} and \citet{nel2015optimal} attempted to show that using the similarity of fundamentals to defined peer groups can increase the precision of the valuation and therefore provide more meaningful  company groupings. \citet{fan2019topology, eaton2022peer} found that while industry classification codes, such as SIC codes, where not of much use by investment bankers when choosing peers, and that the products or services sold by the companies were of much more interest when devising peer groups.
 Recently, some \citep{breitung2023global} have even used advanced large language models (LLMs), a form of NLP, to reveal a company similarity network across most of the world's public companies, and seem to have shown that using such a company similarity network can be used to create peer groups that, when traded, would lead to a significant higher return that using analysts' co-searches, or fundamentals clustering, or industry classification.

\section{Results}\label{results}

Following on the above approaches to finding similar companies based on products \citep{fan2019topology, eaton2022peer} and company descriptions \citep{breitung2023global}, and recognizing that we need a method, e.g., named entity recognition (NER), that can create peer groups in a consistent way with good performance, we decided to study in this paper the efficiency of the so-called Chat Generative Pre-Trained Transformer, commonly known as GPT, in this case the GPT 3.5 (model gpt-3.5-turbo) LLM model \citep{radford2018improving, brown2020language} from OpenAI against a NER base model using spaCy \citep{spacy2}. We researched the ability of both  models in extracting the products and services companies provide based on the publicly available description of those firms  in their respective Wikipedia websites. While LLM models create a statistical probability distribution function of words showing up consecutively, and is based on unsupervised learning\footnote{Unsupervised learning are statistical methods that create a model from data that is not labelled by humans. This is in contrast with supervised learning, where data is labelled, e.g., a set of images together with the label, the description, or to give an example within the subject this paper, one could have the data as the fundamentals of the company, and as the label as the value of the company,  e.g., the dividend yield or the future growth rate \citep[see e.g.,][and reference therein]{berry2019supervised}.}, standard NER models create a neural network, mapping unstructured text to a set of entities based on supervised learning. Standard neural network systems such as spaCy has been, for a while, one of the industry default approaches for NER, however, recently the LLM models have started showing quite a considerable improvement and have become the subject of a social media and news hype.

 To do this work, we first identify 13 publicly listed companies that have a Wikipedia page, and extract the summary part of the Wikipedia page as pure text. The data set is described below in table~\ref{companies_table}.

\begin{table}[!h]
\centering
\centering
\begin{tabular}{@{}lrr@{}}
\toprule
Company Name & Wikipedia Page ID\footnotemark[1]  & Summary length (words) \\
\midrule
Apple Inc.   									& 856 		& 437  \\
International Business Machines Corporation   	& 40379651 	& 452  \\
Iridium Communications Inc.   					& 53008 	& 102  \\
Honda Motor Co., Ltd.   						& 13729 	& 248  \\
Nestl\'{e} S.A.   								& 160227 	& 298  \\
TOTAL SE   										& 804161 	& 127  \\
HSBC Holdings plc   							& 322572 	& 283  \\
BioNTech SE   									& 64671486 	& 215  \\
Garmin Ltd.   									& 1118198 	& 86   \\
ASUSTeK Computer Inc.   						& 43591321 	& 167  \\
Sociedad Qu\'{i}mica y Minera de Chile S.A.   	& 7290045 	& 48   \\
Colliers International Group Inc.   			& 23080364 	& 102  \\
Reliance Power Limited   						& 15232287 	& 179  \\
\bottomrule
\end{tabular}
\footnotetext{Source: This data originates from Wikipedia, sourced on 7 June 2023.}
\footnotetext[1]{The Wikipedia Page ID can be obtained by navigating to the Wikipedia page for the company, then clicking on Tools $\rightarrow$ Page Information.
It can also be done automatically using a computer language script such as python and the Wikipedia API, e.g., one can map a company name to a Wikipedia Page ID, or a Wikipedia Page ID to the actual full URL of the Wikipedia page, among many other mappings.
}
\caption{Data set used from Wikipedia, consisting of 13 companies, from different industries and countries. The exact same data was used for both spaCy based NER and GPT based NER.}
\label{companies_table}
\end{table}

We have kept the above Wikipedia  data set  download frozen as of 7 June 2023, for consistency and to be able to reproduce results. For compatibility with the use of the GPT 3.5 template 
we also remove any page breaks or newlines, and normalize all non-English accentuation, so e.g., ``\ldots Nescaf\'{e}\ldots'' becomes ``\ldots Nescafe\ldots''. We also simply removed any non-ASCII codes\footnote{ASCII stands for American Standard Code for Information Interchange. We used just the basic 95 printable characters.}. A typical example of the company summary is copied below for ``Apple Inc.'', including the products and services which were annotated\footnote{The process of annotation, within the context of NLP, is to assign a type or entity class to a word or small sequence of words, e.g., to assign the entity of place, date, currency to words within a text. Note that we use the same entity -- PRODUCT -- for both product and services, for simplicity.} by hand by us, and we have marked those by bold font for clarity\footnote{For an example of another annotated corpus that can be used for product extraction see e.g., \citet{schon2020corpus}.}. 

\begin{quote}
Apple Inc. is an American multinational technology company headquartered in Cupertino, California. Apple is the world's largest technology company by revenue, with US\$394.3 billion in 2022 revenue. As of March 2023, Apple is the world's biggest company by market capitalization. As of June 2022, Apple is the fourth-largest {\bf personal computer} vendor by unit sales and the second-largest {\bf mobile phone} manufacturer in the world. It is one of the Big Five American information technology companies, alongside Alphabet, Amazon, Meta Platforms, and Microsoft. Apple was founded as Apple Computer Company on April 1, 1976, by  Steve Wozniak, Steve Jobs and Ronald Wayne to develop and sell Wozniak's Apple I {\bf personal computer}. It was incorporated by Jobs and Wozniak as Apple Computer, Inc. in 1977. The company's second {\bf computer}, the Apple II, became a best seller and one of the first mass-produced {\bf microcomputers}. Apple went public in 1980 to instant financial success. The company developed {\bf computers} featuring innovative graphical user interfaces, including the 1984 original {\bf Macintosh}, announced that year in a critically acclaimed advertisement called 1984. By 1985, the high cost of its products, and power struggles between executives, caused problems. Wozniak stepped back from Apple and pursued other ventures, while Jobs resigned and founded NeXT, taking some Apple employees with him. As the market for {\bf personal computers} expanded and evolved throughout the 1990s, Apple lost considerable market share to the lower-priced duopoly of the Microsoft Windows operating system on Intel-powered PC clones (also known as Wintel). In 1997, weeks away from bankruptcy, the company bought NeXT to resolve Apple's unsuccessful operating system strategy and entice Jobs back to the company. Over the next decade, Jobs guided Apple back to profitability through a number of tactics including introducing the {\bf iMac}, {\bf iPod}, {\bf iPhone} and {\bf iPhone} to critical acclaim, launching the Think different campaign and other memorable advertising campaigns, opening the Apple Store retail chain, and acquiring numerous companies to broaden the company's product portfolio. When Jobs resigned in 2011 for health reasons, and died two months later, he was succeeded as CEO by Tim Cook. Apple became the first publicly traded U.S. company to be valued at over \$1 trillion in August 2018, then at \$2 trillion in August 2020, and at \$3 trillion in January 2022. As of April 2023, it was valued at around \$2.6 trillion. The company receives criticism regarding the labor practices of its contractors, its environmental practices, and its business ethics, including anti-competitive practices and materials sourcing. Nevertheless, the company has a large following and enjoys a high level of brand loyalty. It has also been consistently ranked as one of the world's most valuable brands.\footnote{From \href{https://en.wikipedia.org/wiki/Apple_Inc.}{Apple Wikipedia page}, extracted and cached on 7 June 2023. Any spelling or grammar errors were not corrected, neither any attempt was made to make the English (American versus British spelling) consistent, i.e., the texts were used as sourced from Wikipedia.}
\end{quote}

The annotation process is, obviously, quite subjective, it depends on what one considers to be products or services\footnote{There are some publicly available databases of product/services such as the Nice Agreement lists \citep{roberts2012international}
and the United Nations Standard Products and Services Code (UNSPSC) \citep{fairchild2002coding}, but given the small amount of examples to annotate, we decided to just do it by hand.}. However, we tried to be as consistent as possibly within the data set, e.g., assigning both singular and plural versions of words, such as the above ``computer'' and ``computers'' entities. 

For the base standard NER cases, we have annotated the company summaries from Wikipedia and used spaCy models\footnote{We used python spaCy version 3.5.3, the latest as of 7 June 2023, and we have noted and verified that although different version of spaCy and its trained models would give slightly different results,  the qualitative conclusions would remain the same. We have used as a template for our spaCy python code some of the approaches in \citet{Landstein_2020, Landstein}.} with 100 learning steps\footnote{And mini batch size = 1.}, no drop-out\footnote{Drop-out is used in machine learning to avoid the neural network over-fitting the training data, and is implemented by randomly omitting certain neurons at each learning step.} and the neural network optimizer algorithm being the Adam Optimizer \citep{2014arXiv1412.6980K} as it is the default setting in spaCy. We have used the empty English model 'en', and the built-in spaCy 'sm' (small), 'md' (medium) and 'lg' (large) models. These three latter models we have implemented with transfer learning \citep{Yosinski:2014:TFD:2969033.2969197}, that is, we have taken already trained standard models and fine-tuned with extra examples, which is called transfer learning \citep[for an application of transfer learning in another setting, see e.g.,][]{covas2020transfer}. For the GPT model, we used OpenAI's GPT 3.5 (gpt-3.5-turbo model as of 7 June 2023), which has been trained on web data with a cut-off of September 2021. 

Named entity recognition with GPT does not work the same way as in spaCy, since the former is a large language model (LLM), that is, a very large neural network model trained on extremely large set of unstructured texts, in non-supervised way, and in simplified terms, able to predict the next word(s) in a conversation from a prompt or template, while the latter uses supervised learning with a training set that has been labelled or annotated, that is, NER in spaCy is tuned using texts where one tells the neural network what are the types or entities and what words or small sequences of words have those types or entities. Therefore, to further ``train'' GPT 3.5 for our NER product extraction task one must introduce a text format template, so that GPT could recognize that there was a specific format to the input training cases and a specific output format too\footnote{We adapted our GPT text input/output template from the template in the online article published in \citet{Chowdhury_2023}.}. The format of the template can be anything one wants, if it is very clear where the examples are -- so there has to be a clear separation, and a clear fixed format for the product/services list. Below we show an example of this GPT template, using ``Apple Inc.'' as the sole training example:

\begin{ttquotesmall}
Entity Definition:
1. PRODUCT: Short name or full name of product or services sold.\\

\noindent Output Format:
\{\{'PRODUCT': [list of entities present]\}\}
If no entities are presented in any categories keep it None\\

\noindent Examples:\\

\noindent 1. Sentence: Apple Inc. is an American multinational technology company headquartered in Cupertino, California. Apple is the world's largest technology company by revenue, with US\$394.3 billion in 2022 revenue. As of March\ldots\\
\noindent Output: \{\{'PRODUCT': ['mobile phone', 'personal computer', 'computer', 'microcomputers', 'Macintosh', 'iMac', 'iPod', 'iPhone', 'iPad']\}\}
\end{ttquotesmall}

For each example added to the template, we increased the enumeration, from 1 onwards. For inference or prediction rather than training, we just use an empty \texttt{``Output: \{\{'PRODUCT': []\}\}''} product set. We note that the GPT model available in OpenAI always has some randomness in it, and one cannot therefore obtain the same exact output with two identical inputs\footnote{See e.g., the description of this non-determinism behaviour in \url{https://platform.openai.com/docs/guides/gpt/faq}.}. Nonetheless, we have verified that the conclusions derived from the GPT models were robust to this kind of built-in randomness\footnote{As described in OpenAI's chat completion API in \url{https://platform.openai.com/docs/api-reference/completions}.}. In order to be able to assess the performance of GPT versus the base NER models, we used a confusion or error matrix and the F-score \citep[see e.g.,][]{olson2008advanced,powers2011evaluation}, as it is standard in the industry. We first calculated the relevant elements of the confusion matrix, the number of true positives (TP), i.e., the number of predicted product/services that we had in advance classified by hand as real product/services, the number of false positives (FP), i.e., the number of predicted product/services that we had in advance classified as not product/services and the number of false negatives (FN), i.e., the number of in advance classified as product/services that the predicted set did not contain. The precision and recall, and the traditional (or balanced) F-score (which is defined as the harmonic mean of the precision and the recall) can then be formulated \citep[see p. 138 in][]{olson2008advanced} as:

\begin{equation}
\label{equation_precision}
{\text{Precision}} = \frac{\text{TP}}{\text{TP}+\text{FP}},
\end{equation}

\begin{equation}
\label{equation_recall}
{\text{Recall}} = \frac{\text{TP}}{\text{TP}+\text{FN}},
\end{equation}

\begin{equation}
\label{equation_fscore}
{\text{F-score}} = 2 \cdot \frac{\text{Precision} \cdot \text{Recall}}{\text{Precision}+\text{Recall}} = \frac{2 \cdot \text{TP}}{2 \cdot \text{TP}+\text{FP}+\text{FN}},
\end{equation}\\

\noindent
where the precision measures the portion of relevant product/services within the predicted set, and recall the portion of relevant product/services within the benchmark annotated set that were predicted. The F-score then measures the overall performance of the predictors. We notice that we do not consider, for the counting of TP, FP and FN, any case where the entity for the word is ``O'', that is, in NER speak,  a non-entity. This is because, as it is rare to have a product/service word in the text, if we counted a prediction of non-entity, that has been annotated as a non-entity, as a true positive, then the performance would be artificially  high -- one would need just to predict that everything is a non-entity, and that there was no product/services  in the text, to get a high score, which is not what we are aiming for, obviously.

The data set, as standard in machine learning models, is divided into a training and test sets, randomly, across all combinations. The training set within the spaCy models consists of annotated by hand texts while for the GPT model consists of the template  with the text plus a clearly demarcated set of product/services.

We have focused on a very small set of companies\footnote{There is, worldwide, an estimated 328+ million companies (as of 2021) according to \citet{Statista_2023}. Of those, an estimated \SI{58200} are public companies (as of Q1 2022) according to \citet{listed2022}. From those listed companies, not all have an explicit company's page in Wikipedia. By using the company name and the Wikipedia page ID, we have identified at least \SI{3890} public companies with a Wikipedia page, for which we extracted the summary, to be used in the out-of-sample testing and in the building of example peer groups for our (annotated set) of 13 companies.} mainly for 3 reasons. First, and foremost, GPT from OpenAI
 is not a free product, it is a commercial product, with a very limited free trial\footnote{Limited in time, number of runs and number of words/tokens.}. Second, we wanted to try to perform few-shot learning for named entity recognition \citep[see e.g.,][]{huang2021few}, as annotation is expensive and time consuming. Third, GPT from OpenAI, in its current configuration only allows a maximum of 4096  (gpt-3.5) and \SI{16384} (gpt-3.5-turbo-16k) tokens\footnote{The operation of tokenization transforms a sentence or text into a set or list of tokens, demarcations of groups of characters, usually words.} or words for each run, and this seriously impairs the ability to run a larger number of training examples. Fourth, and finally, the results, as we shall see later, were already quite good with a few training examples. So, we have first trained/tested on 13 fully annotated examples, and then used a much larger list of around \SI{3890} publicly listed companies for which we could find a Wikipedia Page ID and therefore a Wikipedia page with a summary text\footnote{We note that many of these companies Wikipedia pages have extra structured information, such as the industry classification, the main exchange ticker, even the product/services themselves. However, not all have them, and the format can vary a lot, making it very difficult to extract these data elements uniformly. The summary, by contract, is present in almost all companies' Wikipedia websites.}. 

The first result we have noticed is that even with no examples (zero examples in the template, so a case of the so-called zero-shot learning \citep[see e.g.,][]{wei2021finetuned}), GPT was able, straight out of the ``box'', to predict quite a few product/services. As an example, for the company ``Sociedad Qu\'{i}mica y Minera de Chile S.A'', we can see below that it predicts quite well (below, the ``set product benchmark'' is our own annotation, while ``predicted set product from GPT'' are the results from GPT 3.5).

\begin{ttquotesmall}
\noindent Sociedad Quimica y Minera de Chile (SQM) is a Chilean  chemical company and a supplier of plant nutrients, iodine, lithium and industrial chemicals. It is the world's biggest lithium producer.SQM's natural resources and its main production facilities are located in the Atacama Desert in Tarapaca and Antofagasta regions.\\

\noindent set product benchmark ['iodine', 'plant nutrients', 'chemical', 'industrial chemicals', 'lithium']\\

\noindent predicted set product from GPT \{\{'PRODUCT': ['plant nutrients', 'iodine', 'lithium', 'industrial chemicals']\}\}\\

\noindent f\_score=0.88
\end{ttquotesmall}

The second result was that GPT does not seem to confuse the product / services that the company has with the market for what the product / services is meant for. For example, for the company ``Garmin Ltd'', on a GPT run with 3 training examples, it extracted the following words as product/services:
\begin{ttquotesmall}
\ldots Schaffhausen, Switzerland.The company specializes in {\bf GPS technology} for automotive, aviation, marine, outdoor, and sport activities. Due to their development \ldots
\end{ttquotesmall}
It marked the ``GPS technology'' as the correct annotated product/service, and ignored correctly the words 
``automotive'', ``aviation'', ``marine'', ``outdoor'', and ``sport''. We suspect that it understands (via its statistical probability model) that the word ``for'' means the ``market of'', rather than the product/service itself. This was quite impressive compared to the base spaCy model, which can return e.g., ``aviation'', ``activity tracker'', ``marine'', ``GPS technology'', ``automotive'', 'smartwatch consumer'', getting confused between product/services and their respective markets.

A further result we were surprised to see was that GPT was able to disambiguate some text such as ``\ldots property and asset management \ldots'' (for company ``Colliers'') into two product/services: ``property management'', and ``asset management''. This would be quite impressive  for a human  to do it, but for GPT to be able to do this, although not in every single run/example, was quite surprising. 

Our main result is the direct comparison of the performance of the base spaCy models, representing a standard form of doing NER, with the GPT LLM models. We have run the empty English model 'en', and the built-in spaCy 'sm' (small), 'md' (medium) and 'lg' (large) models, against  OpenAI's GPT (gpt-3.5-turbo) for the number of training examples from 0 to 9. The results of the F-score, our standard measure of performance, are depicted below in Figure~\ref{comparison}.

\begin{figure}[htp]
\centering
\includegraphics[width=10cm]{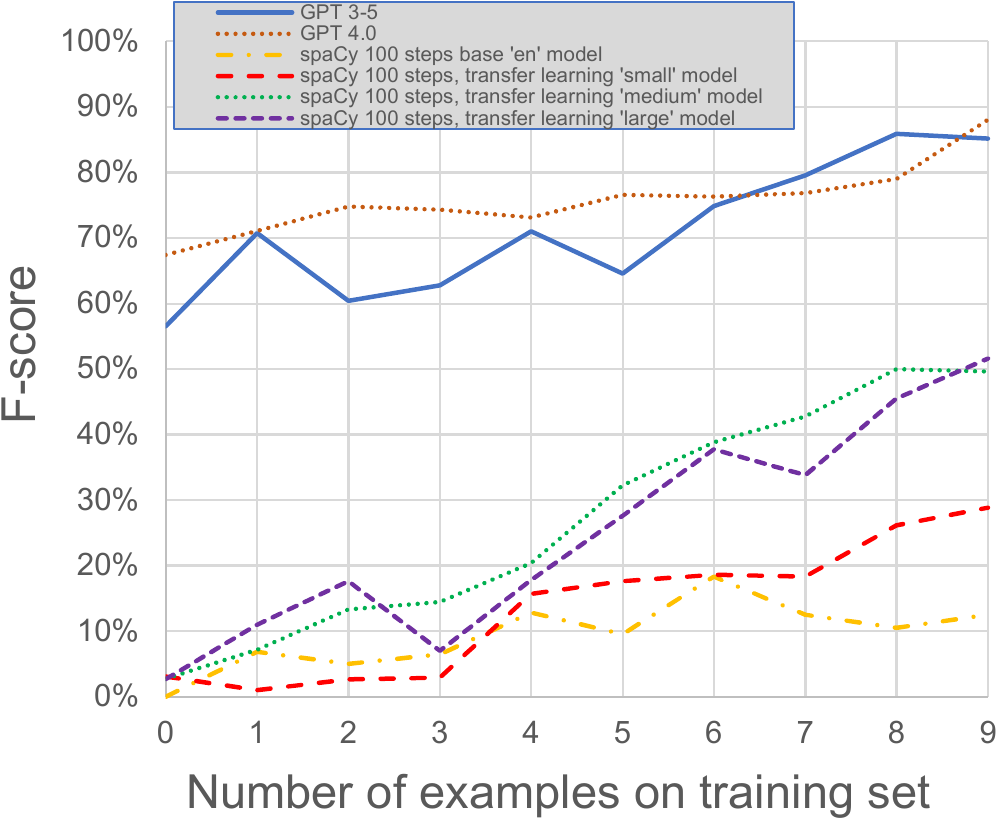} 
\caption{Comparison of F-score statistic showing product/services extraction performance using named entity recognition (NER), for GPT 3.5 versus standard NLP-NER models. Notice that for zero training examples (zero-shot learning), the GPT F-score is already $56.51\%$, much higher than the standard models. In particular, the F-score zero-shot learning for the spaCy empty English model 'en' is $0\%$, as it is expected. On the other end of the scale, with nine training examples, GPT already achieves a F-score of $85.15\%$, which is quite remarkable given the small amount of training examples. {\bf Note:} The chart has been updated with data for the new gpt-4 model, which shows a small but consistent improvement over gpt-3.5, and is able to reach a F-score of $88.05\%$.
}
\label{comparison}
\end{figure}

The results show that the GPT model clearly outperforms the standard spaCy models, by quite a large margin, consistently, across all the parameter space depicted, i.e., the number of examples within the training set. The best averaged result was around $\text{F-score}=85\%$ for the GPT model, which indicates that the GPT model  can probably already be used for real world cases, that is, to obtain comparable companies peer groups in equity valuation. We have stress tested both the GPT and the spaCy models extensively, and concluded that both models' results were quite robust, with the value of the F-score barely changing for different parameter model changes\footnote{For the GPT model we have stress tested using lower case for all text and product/services, changing the random seed, changing the ``temperature'' parameter which controls the randomness of GPT output, fine tuning the parameters ``max\_tokens'', ``top\_p'', ``presence\_penalty'' and ``frequency\_penalty'' \citep[for the GPT OpenAI parameter API reference see][]{openAI}. We also tested adding empty texts with empty product/services sets to avoid the so-called LLM hallucination problem \citep{lee2023mathematical} but saw no clear improvement.
For the spaCy models we have stressed tested also several parameters, such as using lower case for all text and product/services, the number of iterations of the learning algorithm, the ``dropout'' rate, which allows for some randomness and generalization, the optimization algorithm itself, the parameters of the Adam algorithm such as 
``learn\_rate'', ``beta1'', ``beta2'', ``eps'', ``L2'', ``L2\_is\_weight\_decay'', ``grad\_clip'' \citep[see spaCy API reference in][]{spaCy}.}.

In order to be able to use the results for constructing company's comparable or peer groups, one more step was needed, that is, to use a much larger set of out-of-sample data\footnote{Out-of-sample data is a set of data the machine learning models have not seen, and are not annotated, so on that data we can only do a prediction using the trained model on what the product/services should be.} and find which companies are closest, in the sense of having the highest number of same product / services. This data set is described below in table~\ref{out_of_sample}.

\begin{table}[!h]
\centering
\begin{tabular}{@{}lrr@{}}
\toprule
Out-of-sample companies data set statistics\\
\midrule
Number of public companies with Wikipedia page 						& 3890   \\
Number of above companies that had a summary page 						& 3887   \\
Number of above companies that GPT could extract at least one product/service 						& 3800   \\
Average number of words or tokens on each summary page   		& 108   \\
\bottomrule
\end{tabular}
\footnotetext{Source: This data originates from Wikipedia, sourced on 23 June 2023.}
\caption{Data set statistics for set used for out-of-sample prediction.}
\label{out_of_sample}
\end{table} 

In order to obtain comparable companies for the companies in our annotated set, first we trained GPT with all 13 companies in the training template\footnote{We note that for gpt-3.5-turbo the maximum number of tokens is exceeded when using all 13 companies in the annotated set. Therefore, we had no choice but to use gpt-3.5-turbo-16k which is even more expensive that gpt-3.5-turbo but has the same LLM model and allows four times the number of tokens.}. We have decided, given the cost of GPT and the results we had already obtained, i.e., that most extra parameters do not seem to improve the performance, to just use GPT (gpt-3.5-turbo-16k) with its default settings. Below we show a couple of examples of what the results can be.

\begin{table}[!h]
\centering
\begin{tabular}{@{}lrlr@{}}
\toprule
Comparable company 					& Page ID & Product/services match & Count \\ 
\midrule
Chevron Corporation 						& 284749 	& ['gas', 'energy', 'power generation', 'refining', 'oil'] 	& 5 \\
China Petroleum \& \ldots \ldots    		 & 1325529 	& ['gas', 'natural gas', 'refining', 'oil', 'crude oil'] 	& 5 \\
Sinopec Oilfield  \ldots 		& 1325529   & ['natural gas', 'gas', 'crude oil', 'oil'] 				& 4 \\
Pilipinas Shell  \ldots 		& 23409980 	& ['gas', 'power generation', 'chemicals', 'oil'] 			& 4 \\
BP p.l.c. 									& 18998720 	& ['refining', 'gas', 'oil', 'power generation'] 			& 4 \\
NextEra Energy, Inc. 						& 28014927  & ['natural gas', 'oil', 'energy'] 							& 3 \\
Indian Oil  \ldots 				& 47428211 	& ['natural gas', 'refining', 'oil'] 						& 3 \\
Hess Corporation 							& 2086678	& ['natural gas', 'energy', 'crude oil'] 				    & 3 \\
BP Prudhoe Bay  \ldots 				& 15366815 	& ['natural gas', 'oil', 'crude oil'] 						& 3 \\
Santos Limited 								& 1140903 	& ['natural gas', 'gas', 'oil'] 							& 3 \\
\bottomrule
\end{tabular}
\footnotetext{Source: This data originates from Wikipedia, sourced on 23 June 2023.}
\caption{Prediction for one of the companies in the training annotated set (TOTAL SE) with Wikipedia Page ID 804161 and annotated set of products being the set  ['energy', 'petroleum', 'oil', 'gas', 'crude oil', 'natural gas', 'power generation', 'refining', 'petroleum product', 'crude oil trading', 'chemicals'].}
\label{out_of_sample1}
\end{table} 

On table~\ref{out_of_sample1} we can see that the results seem, qualitatively, quite good. All companies identified as within the peer group are energy companies, with similar product/services.

\begin{table}[!h]
\centering
\begin{tabular}{@{}lrlr@{}}
\toprule
Comparable company 					& Wikipedia Page ID & Product/services match & Count \\ 
\midrule
UrtheCast Corp. 						& 32320197 	& ['satellites'] 			&	1 		 \\
Mitsubishi Electric Corporation 		& 1075261 	& ['satellites'] 			& 	1 		 \\
China Satellite Communications Co. 		& 54647879  & ['satellites'] 			& 	1 		 \\
Raytheon Technologies Corporation 		& 63554945 	& ['satellites'] 			&  1 		 \\
The Boeing Company 					    & 18933266  & ['satellites'] 			&  1 		 \\
Hexcel Corporation 						& 8207797 	& ['satellites'] 			& 	1 		 \\
China Unicom (Hong Kong) Limited 		& 451805 	& ['data communication']	& 1 	     \\
SES S.A. 								& 2415553 	& ['satellites'] 			& 1 	     \\
NTT DOCOMO, INC. 						& 374039    & ['satellite'] 			& 1          \\
Semtech Corporation                     & 8484883   & ['satellites'] 			& 1			 \\
\bottomrule 
\end{tabular}
\footnotetext{Source: This data originates from Wikipedia, sourced on 23 June 2023.}
\caption{Prediction for one of the companies in the training annotated set (Iridium Communications Inc.) with Wikipedia Page ID 53008 and annotated set of products being the following set  ['satellite', 'satellites', 'voice communication', 'data communication', 'handheld satellite phones', 'satellite messenger communication devices', 'integrated transceivers', 'two-way satellite messaging service', 'Android smartphones'].}
\label{out_of_sample2}
\end{table}
 
On table~\ref{out_of_sample2} we can see what happens when a company with a more niche market, such as Iridium, which trades on satellite communications is used as a target company for which to find a peer group. The number of word matches is much lower (mostly related to the word/token ``satellite'') and therefore the quality decreases. Nonetheless, it is still qualitatively a reasonable peer group\footnote{We note that our method of selecting the number or count of product/services that two companies share can be used to create a company network, as described in \citet{fan2019topology}. In their paper, they build a company network based on the similarity of products using filing (texts) of USA listed firms in SEC. They claim that the probability distribution function of the edge (links) strength between companies follows an exponential law. However, we found that in our case a power law $s \propto k ^{-\gamma}$, where $s$ is the strength or count of common product/services, was a better fit.}.
Therefore, the results seem to show that  this method could be used for creating companies' comparable peer groups, and that the results seem reasonably good.

We have used Wikipedia data as we wanted to use exclusively publicly available web-based data. However, if one would could use the data sets from commercial data providers such as S\&P, Bloomberg, FactSet, Reuters or others \citep[see e.g.,][and reference therein]{jha2019implementing}, it would allow to build a company's peer group automatically for all public (and private companies). Therefore, this would then industrialize one of the last elements of a company valuation pipeline to be automated. We have used \SI{3890} companies, which was what we could find manually on Wikipedia, but we note that commercial data providers have clean, curated and structured data for all {\SI{58200}+} publicly listed  companies and a large number of the private companies as well. We also emphasise that doing the data analysis on all public companies and implementing a full peer group selection using GPT would imply quite a monetary cost, likely to only be able to be sponsored by a commercial fintech or investment banking house, as the usage would exceed easily what can be done with research based GPT free trials and research budgets.

Finally, we note that in the future we plan to use a company valuation model from e.g., a software house specializing in valuation, to be able to prove that the product/services obtained from GPT text-based NER can be used to create a better company peer groups with more accurate valuations (e.g., as compared with the average stock market price derived valuation). However, given the number of companies worldwide, any kind of large-scale usage of our approach would require some investment, as GPT from OpenAI is currently not free and is therefore expensive for any larger than small research projects utilization.

\section{Conclusion}\label{conclusion}

Using companies' description/summaries from publicly available Wikipedia data, we have shown quantitatively that using large language models (LLMs) such as GPT, results in a much higher performance and success rate than standard named entity recognition (NER) which uses manual annotation and systems such as spaCy. We have shown this in the specific case of product/services entity extraction. Furthermore, we shown,  qualitatively, that this entity extraction by GPT models can be used to create companies' peer groups that look reasonable and consistent. This is suggestive that these LLM models could, in the future, be used for helping the  automation of companies' peer group construction, and therefore the full automation of the company valuation pipeline.

\section*{Declarations}

\begin{itemize}
\item Funding: The author and this research was fully self-funded by the author.
\item Conflict of interest/Competing interests: None
\item Availability of data and materials: All data sets used were from publicly available websites on the Internet (including Wikipedia) and their sources are referenced/cited in the text.
\end{itemize}

\bibliography{my_bibliography.bib}

\end{document}